\documentclass{article}

\usepackage{arxiv}

\usepackage[utf8]{inputenc} 
\usepackage[T1]{fontenc}    
\usepackage{hyperref}       
\usepackage{url}            
\usepackage{booktabs}       
\usepackage{amsfonts}       
\usepackage{nicefrac}       
\usepackage{microtype}      
\usepackage{lipsum}
\usepackage{graphicx}
\graphicspath{ {./images/} }
\usepackage{xcolor}
\usepackage[table]{xcolor}
\usepackage{booktabs}
\usepackage{makecell}
\usepackage{amsmath}
\usepackage{longtable}
\usepackage{amsmath}
\usepackage{bibentry}
\usepackage{hyperref}

\title{Fusion Embedding for Pose-Guided Person Image Synthesis with Diffusion Model
\thanks{\textit{This article was published in ACM Transactions on Multimedia Computing, Communications, and Applications (ACM TOMM) on April 12, 2026.}}}

\author{
 Donghwan Lee \\
  Department of Industrial Engineering\\
  Yonsei University\\
  Seoul, Republic of Korea \\
  \texttt{dhlee.ie@yonsei.ac.kr} \\
  \And
 Kirok Kim \\
  Department of Industrial Engineering\\
  Yonsei University\\
  Seoul, Republic of Korea \\
  \texttt{alfmalfm11@yonsei.ac.kr} \\
  \And
 Jisu Lee \\
  Department of Industrial Engineering\\
  Yonsei University\\
  Seoul, Republic of Korea \\
  \texttt{lejss1012@yonsei.ac.kr} \\
  \And
 Kyungha Min \\
  Department of Industrial Engineering\\
  Yonsei University\\
  Seoul, Republic of Korea \\
  \texttt{0825pm@gmail.com} \\
  \And
 Wooju Kim\thanks{Corresponding author} \\
  Department of Industrial Engineering\\
  Yonsei University\\
  Seoul, Republic of Korea \\
  \texttt{wkim@yonsei.ac.kr} \\
}

\renewcommand{\today}{}

\begin{document}
\maketitle

\begin{abstract}
Pose-Guided Person Image Synthesis (PGPIS) aims to generate human images in specified poses while preserving the identity and appearance of a source image. This technology facilitates diverse applications, including virtual try-on, digital avatars, animation, and sign language generation. Despite the high-quality results of recent diffusion-based PGPIS, these models typically depend on implicit feature aggregation within the denoising process. As a result, fine-grained texture preservation is limited, and even for the same identity, it is difficult to ensure consistent generation under variations in pose and source appearance. To address these limitations, we propose Fusion Embedding for PGPIS using a Diffusion Model (FPDM), the first framework that explicitly aligns fused source–pose embeddings with target image embeddings via contrastive learning, and subsequently employs the learned fusion embedding as a conditioning signal for generation. FPDM integrates an Image–Pose Fusion (IPF) module into our proposed Source-Enhanced Pose Fusion approach to learn a fusion embedding aligned with the target image. We then employ a conditional diffusion model guided by source appearance, target pose, and the learned fusion embedding. Experiments on the DeepFashion benchmark and the RWTH-PHOENIX-Weather 2014T dataset demonstrate competitive performance compared to existing methods in both quantitative and qualitative evaluations, with ablation studies confirming that explicit fusion embedding alignment substantially improves texture fidelity and consistency across pose and source appearance variations. The implementation is publicly available at https://github.com/dhlee-work/FPDM.
\end{abstract}


\section{Introduction}
Generating realistic virtual human images is an important research problem in computer vision, as it supports applications such as virtual reality, gaming, e-commerce, and multimedia content creation. Among various approaches, Pose-Guided Person Image Synthesis (PGPIS) has attracted significant attention. PGPIS aims to generate a photo-realistic image of a person in a specified target pose while preserving appearance attributes such as clothing, identity, and texture from a given source image \cite{RN6}. These capabilities are particularly useful for virtual try-on systems, avatar-based interactions, and sign language video generation for the hearing-impaired \cite{RN1}.

Early research on PGPIS was dominated by Generative Adversarial Networks (GANs) \cite{RN2,RN23,RN19,RN24,RN25,RN26,RN27,RN207}. While GANs generate realistic images, unstable training and single-pass synthesis limit their ability to preserve fine details. This typically manifests as blurry contours or distorted textures in the resulting images \cite{RN67,RN23}.

Diffusion models (DMs) \cite{RN37} have recently emerged as a powerful alternative, achieving state-of-the-art results in image synthesis. Conditional diffusion models \cite{RN47,RN95} incorporate guidance signals to control generation, and several diffusion-based PGPIS methods \cite{RN30,RN38,RN33,RN36,RN211,RN34,RN209,RN208,RN206} leverage both the source image and the target pose as conditions. By iteratively refining noisy latents, diffusion models can better capture structural details. While the first PGPIS diffusion model, PIDM \cite{RN30}, demonstrated feasibility in pixel space, subsequent methods have shifted toward Latent Diffusion Models (LDMs) \cite{RN49}. This transition aims to reduce computational costs and facilitate high-resolution image generation. 

\begin{figure*}[t]
\centering
\includegraphics[width=0.9\textwidth]{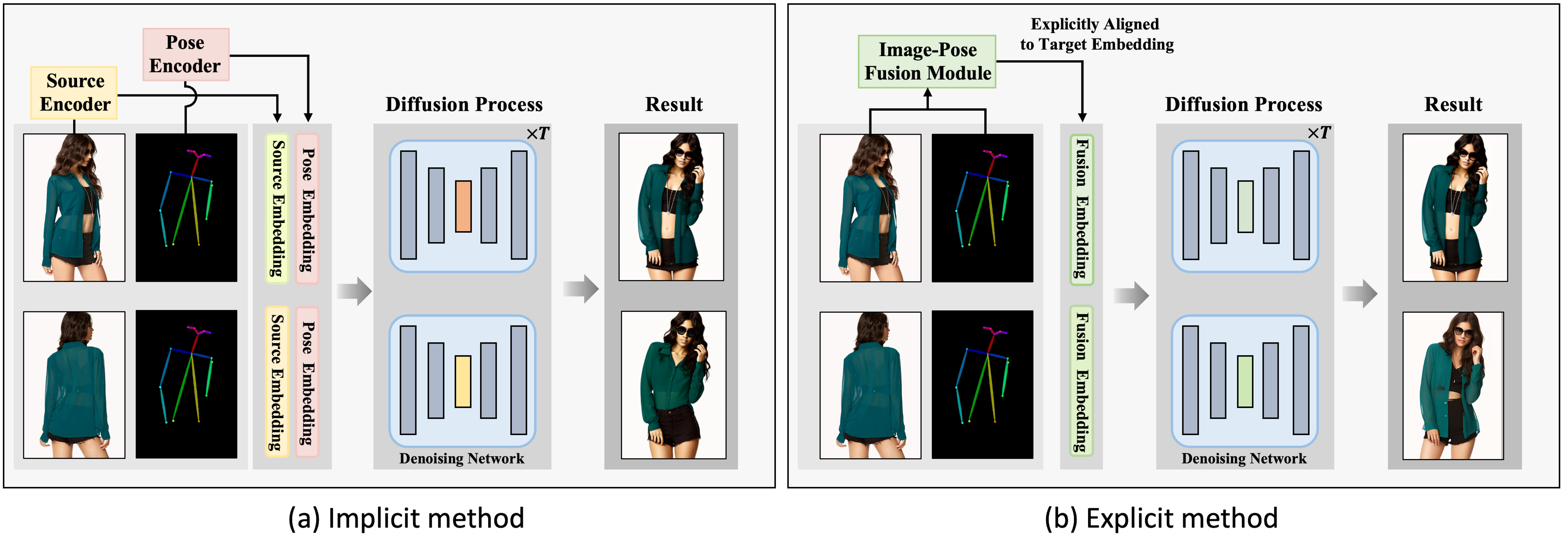} 
\caption{Conceptual comparison of conditioning strategies in pose-guided person image synthesis. (a) Implicit Method: Source appearance and target pose embeddings are directly injected into the denoising network at each diffusion timestep, and their interaction is learned implicitly during training. Throughout the generation process, source appearance and target pose conditioning become repeatedly entangled within the denoising network, resulting in inconsistent generation outcomes for the same identity under different configurations of source appearance and target pose conditions. A representative case with a fixed target pose is shown (e.g., outputs from MCLD\cite{RN209}). (b) Explicit Method (Ours): Our framework introduces an explicit Image–Pose Fusion module that combines source appearance and target pose embeddings to construct a fused representation, which is aligned with the target image embedding via contrastive learning prior to diffusion. This pre-aligned representation provides a stable conditioning signal, enabling robust identity preservation and consistent pose synthesis. Representative results generated by the proposed method under the same target pose condition are shown.}
\label{fig:fig0}
\end{figure*}

Despite these advances, preserving fine-grained appearance details and achieving consistent generation for the same identity under variations in pose and source appearance remain challenging in diffusion-based PGPIS. This difficulty largely stems from the reliance on implicit conditioning in most existing approaches, where source appearance and target pose are incorporated into the diffusion process without explicit alignment or structuring beforehand. As illustrated in Figure~\ref{fig:fig0}(a), source appearance and target pose embeddings are injected into the denoising network (e.g., pose into the noise input and appearance into internal UNet layers), and their interaction is learned implicitly by the denoising network during training. As a result, the relationship between appearance and pose becomes repeatedly entangled within the network, leading to a denoising-process-dependent interaction between source appearance and pose. In terms of generation results, such entanglement often leads to inconsistent outcomes, even for the same identity, when only the source appearance varies under a fixed target pose, manifesting as inconsistencies in visual appearance and identity-related details. This observation suggests that pose-consistent generation requires a conditioning representation that is explicitly aligned and invariant to source appearance variations before diffusion begins.

To address this limitation, we propose FPDM (Fusion Embedding for PGPIS with Diffusion Model).
FPDM introduces a dedicated contrastive learning approach in a pre-diffusion stage to explicitly align
a fused image–pose representation, derived from the source image and target pose, with the target image embedding. The resulting aligned fusion representation is used as a stable conditioning signal throughout the denoising process, enabling consistent image generation despite variations in the source image or target pose. As illustrated in Figure~\ref{fig:fig0}(b), the explicitly aligned image–pose fusion embedding enables stable and appearance-invariant generation with faithful identity preservation. FPDM consists of two primary components: (i) an Image-Pose Fusion (IPF) module, which utilizes a Combiner Network \cite{RN45} and Source-Enhanced Pose Fusion contrastive learning to align source-pose embeddings with the target image; and (ii) a conditional diffusion model guided by three inputs: DINOv2 source patch embeddings, CNN-based target pose features, and the IPF fused embedding. Extensive experiments on DeepFashion and RWTH-PHOENIX-Weather 2014T demonstrate that FPDM achieves competitive performance to state-of-the-art methods in both quantitative and qualitative evaluations. 
Moreover, ablation studies confirm that the fusion embedding enhances generation quality in terms of texture fidelity and improves robustness to pose and appearance variations for identity preservation. While the proposed framework can be conceptually extended to other video generation settings, this work focuses on pose-guided person image synthesis (PGPIS), where strict identity preservation under large pose variations is essential. Accordingly, we do not position our method as a general-purpose video generation model, but rather as a task-specific approach tailored to PGPIS and its direct applications, such as sign language video generation. The main contributions of this work are as follows:

\begin{itemize} 
\item We propose FPDM, the first diffusion-based PGPIS framework with contrastive fusion embedding, enabling explicit alignment of source appearance and target pose. 
\item We design an Image‑Pose Fusion (IPF) module based on the Source‑Enhanced Pose Fusion Approach, which leverages a Combiner Network to fuse source and pose embeddings via contrastive learning, enabling robust conditional synthesis.
\item We demonstrate state-of-the-art performance on two benchmark datasets. FPDM provides strong texture preservation and visual consistency, benefiting multimedia applications such as sign language video generation and avatar synthesis.
\end{itemize}

\section{Related Work}

\subsection{Diffusion Models for Image Synthesis}
 Diffusion models (DMs) \cite{RN37} have recently achieved remarkable success in high-quality image synthesis. They are trained via a forward process, which gradually adds noise to data, and a reverse denoising process that reconstructs the image step by step. Conditional diffusion models extend this framework by incorporating guidance signals to control generation. Classifier-guided diffusion \cite{RN47} and classifier-free guidance \cite{RN95} are two representative methods for conditional synthesis. To reduce computational costs and support high-resolution outputs, the Latent Diffusion Model (LDM) \cite{RN49} performs diffusion in the latent space of a variational autoencoder. Based on this approach, Stable Diffusion (SD) \cite{RN50} has become a widely used model by combining latent diffusion with CLIP text embeddings. These advances have established diffusion as a powerful backbone for conditional generation tasks, including pose-guided image synthesis.

\subsection{Contrastive Learning for Conditioning}
Contrastive learning (CL) learns representations by bringing similar pairs closer and pushing dissimilar ones apart in the embedding space. It possesses strong generalization capabilities, does not rely on explicit labels, and has been widely adopted in foundation models \cite{RN200}. CLIP \cite{RN41} aligns image–text pairs through contrastive learning and is extensively used in vision–language tasks and generative models such as Stable Diffusion \cite{RN50}. DINO \cite{RN201}, based solely on image data, performs self-distillation with contrastive learning and produces features with strong transferability to various vision tasks. CLIP4CIR \cite{RN45} proposed composed image retrieval by introducing a Combiner network that produces a fused image–text embedding explicitly aligned with the target image embedding. This paradigm is particularly relevant to Pose-Guided Person Image Synthesis (PGPIS). PGPIS involves combining heterogeneous embeddings—source appearance and target pose—into a coherent target image. While prior diffusion-based methods relied on implicit alignment to bridge these modalities, contrastive learning enables a more rigorous, explicit alignment. 

\subsection{Pose-Guided Person Image Synthesis} 
PGPIS aims to generate a realistic image of a person in a target pose while preserving the appearance of a given source image \cite{RN2}. Early research primarily used GAN-based models, but their min–max training often caused instability and limited fine-grained detail preservation. Recently, diffusion-based PGPIS has emerged as a promising alternative. The first such method, PIDM \cite{RN30}, utilized a CNN to extract conditional features from the pose image and applied them to guide a pixel-level diffusion process. Subsequent approaches have leveraged Latent Diffusion Models (LDMs) to reduce computational overhead and improve scalability for high-resolution generation. To better exploit source appearance, several strategies have been explored, including learnable query-based refinement (CFLD \cite{RN33}), body-part or region-level embedding extraction (DRDM \cite{RN211}, MCLD \cite{RN209}), and attention or flow-based alignment (Leffa \cite{RN206}). Another research line incorporates both source and target pose alignment, including pose‑constrained attention (PoCoLD \cite{RN38}), recurrent alignment with a Mapformer (RePoseDM \cite{RN34}), Transformer‑based condition aggregation (XMDPT \cite{RN93}), inpainting‑based multi‑pose generation (ImagePos \cite{RN208}), and a three‑stage Prior–Inpainting–Refinement framework (PCDM \cite{RN36}). Although PCDM attempts to predict global embeddings in a prior stage, it primarily relies on regression-based objectives (e.g., MSE), which may overlook the intricate semantic interdependencies between source appearance and target pose. In contrast, FPDM leverages a contrastive learning-based approach to explicitly align the fused image-pose representation with the target image embedding within a shared latent space. This strategy yields a more discriminative and pose-consistent conditioning signal than traditional structural mapping or regression-based estimation. In contrast to these approaches, recent efforts have begun to explore explicit alignment strategies between fused conditional representations and target images using contrastive objectives.

\subsection{Conditional video generation}
The success of diffusion-based image synthesis has extended to video generation research, where conditional video generation has emerged as an important topic in recent years. Conditional video generation has been actively studied across a variety of applications, including text-based video generation\cite{RN212,RN219,RN220}, pose-conditioned video generation\cite{RN213, RN221, RN223}, and video inpainting\cite{RN218,RN217,RN224}. In particular, in the context of addressing recent societal challenges, sign language video generation\cite{RN64,RN204,RN216,RN215,RN205,RN214} has received increasing attention. This task requires effective relational learning between textual or pose-based inputs and spatiotemporal motion representations. To this end, recent studies have explored diffusion models. For instance, SignDiff \cite{RN205} employs a dual-condition diffusion model with a Frame Reinforcement Network to bridge the gap between skeletal poses and appearance, surpassing earlier GAN-based methods in temporal coherence. More recently, SignViP \cite{RN214} introduces a discrete tokenization strategy to integrate fine-grained conditions like 3D hand representations, further improving fidelity through a multi-condition token space. Most existing diffusion-based approaches for pose-conditioned sign language video generation implicitly model interactions between pose and source appearance conditioning signals, making them relevant reference cases for our work. Accordingly, we conduct additional evaluations on sign language video datasets to examine the practical applicability of the proposed method.

\section{Method}

\subsection{Preliminary}
The Stable Diffusion (SD) model \cite{RN50} is built upon the Latent Diffusion Model (LDM) \cite{RN49}, which consists of a Variational Autoencoder (VAE) \cite{RN74} for encoding images into a low-dimensional latent space and a UNet \cite{RN73} for denoising noisy latents. SD adopts the Denoising Diffusion Probabilistic Model (DDPM), which consists of a forward diffusion process and a reverse denoising process. In the forward process, a noisy latent $z_t$ at timestep $t \in [1,T]$ (typically $T=1000$) is obtained by adding Gaussian noise $\epsilon \sim \mathcal{N}(0,1)$ to the initial latent $z_0$:

\begin{equation}
    z_t= \sqrt{\bar\alpha_t}\ z_0 + \sqrt{1-\bar\alpha_t }\ \epsilon,
    \label{eq:eq1}
\end{equation}

where $\bar\alpha_t=\prod_{s=1}^{t}\alpha_s$ is the cumulative product of the noise schedule. As $t$ increases, the latent variable $z_t$ gradually approaches a Gaussian noise distribution.

In the reverse process, the model predicts the noise in $z_t$ using a conditional UNet $\epsilon_\theta(z_t,t,c)$, where $c$ denotes a conditioning embedding (e.g., a text embedding from CLIP \cite{RN41}). The training objective is defined as the mean squared error between the true noise and the predicted noise:

\begin{equation}
    \mathcal{L} = \mathbb{E}_{z_0,c,\epsilon,t}\left[ \left\| \epsilon - \epsilon_{\theta} (z_t, t, c) \right\|_2^2 \right].
    \label{eq:eq2}
\end{equation}

During image generation, the process starts from Gaussian noise $z_{T} \sim \mathcal{N}(0,I)$ and iteratively denoises it using the trained $\epsilon_\theta$ to generate an image consistent with the conditioning $c$.

\subsection{Image-Pose Fusion Module}    

\begin{figure*}[t]
\centering
\includegraphics[width=0.9\textwidth]{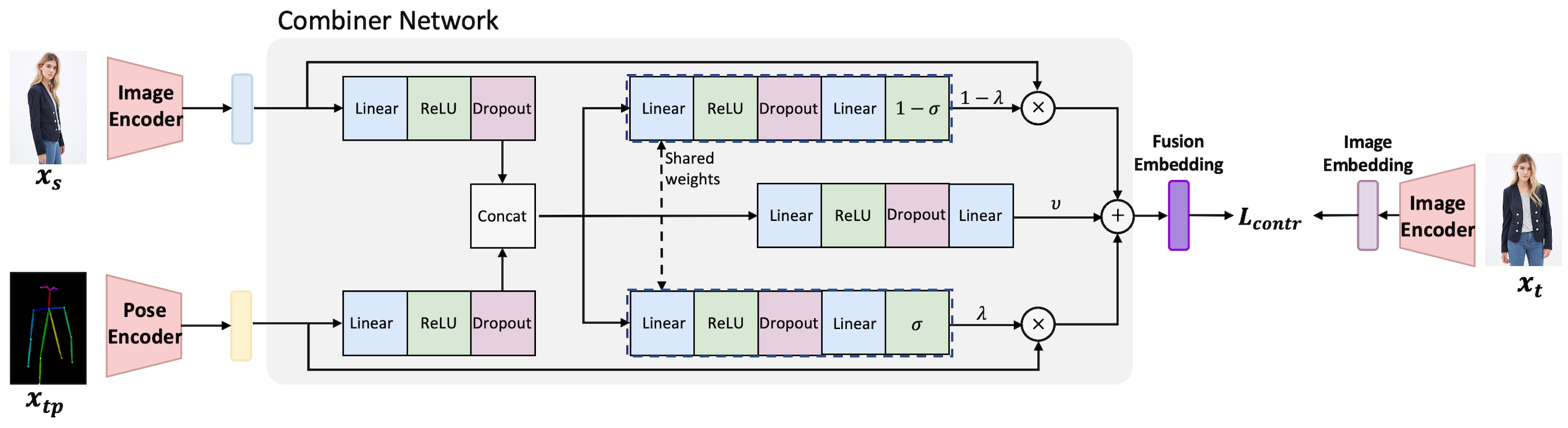} 
\caption{Architecture of the Image-Pose Fusion (IPF) module for aligning source image and target pose embeddings.}
\label{fig:fig1}
\end{figure*}

The Image-Pose Fusion (IPF) module is designed to produce a fused source–pose embedding that is explicitly aligned with the target image representation via contrastive learning, thereby ensuring consistent image synthesis.

Figure~\ref{fig:fig1} provides an overview of the IPF module architecture. The module consists of three components: an Image Encoder (IE), a Pose Encoder (PE), and a Combiner (Comb). The CLIP-based Image Encoder extracts embeddings from the source image $x_s$ and the target image $x_t$, while the CLIP-based Pose Encoder extracts an embedding from the target pose image $x_{tp}$. These embeddings are fused by the Combiner module~\cite{RN45} to produce a unified representation $e_f = \mathrm{Comb}(e_s, e_{tp})$, where $e_s = \mathrm{IE}(x_s)$ and $e_{tp} = \mathrm{PE}(x_{tp})$. The Combiner integrates source appearance and target pose features through concatenation followed by nonlinear transformations, and adaptively weights each input to capture complementary information. The resulting fused embedding $e_f$ is trained to align with the target image embedding $e_t$ using InfoNCE loss~\cite{RN96}.

During training, each batch consists of triplets $\{x_s, x_{tp}, x_t\}$ corresponding to the same identity. For each triplet, the IPF module produces a fused embedding–target embedding pair $(e_f, e_t)$, which is treated as a positive pair in the contrastive objective. Target embeddings from other triplets within the same batch are used as negative samples, which typically correspond to different identities.

Finally, we compare our Source-Enhanced Pose Fusion approach with a baseline contrastive learning approach. Unlike our method, the baseline does not explicitly distinguish between source and target image embeddings during fusion, which limits its ability to disentangle appearance and pose information.

\subsubsection{Baseline Approach}
The baseline approach learns a fused representation by directly combining the source image and the target pose into a single embedding $e_f$, which is optimized to align with the corresponding target image embedding $e_t$ using a batch-based InfoNCE contrastive loss. Unlike the Source-Enhanced Pose Fusion Approach, this baseline does not explicitly separate source appearance and target pose information during fusion. 

During training, each batch consists of triplets $\{x_s, x_{tp}, x_t\}$ corresponding to the same identity. For each triplet, the model produces a fused embedding–target embedding pair $(e_f, e_t)$, which is treated as a positive pair in the contrastive objective. Target image embeddings from other triplets within the same batch are used as negative samples. As a result, the learning objective encourages $e_f$ to be close to its matched target embedding while being well separated from target embeddings of other triplets in the batch. Following~\cite{RN45}, the batch-based InfoNCE contrastive loss is defined as:

\begin{equation}
    \mathcal{L}_{\text{contr}}
    = \frac{1}{B} \sum_{i=1}^{B} 
    \left[ 
      -\log 
      \frac{\exp\big(\tau \cdot d(\psi_f^i, \psi_t^i)\big)}
           {\sum_{j=1}^{B} \exp\big(\tau \cdot d(\psi_f^i, \psi_t^j)\big)}
    \right],
    \label{eq:eq3}
\end{equation}

where $\psi_f^i$ and $\psi_t^i$ denote the fused embedding $e_f$ and the target image embedding $e_t$ of the $i$-th sample in the batch, respectively. The function $d(\cdot,\cdot)$ represents cosine similarity, $\tau$ is a temperature parameter controlling the sharpness of the logits, and $B$ denotes the batch size.

\subsubsection{Source-Enhanced Pose Fusion Approach} 
While the Baseline Approach aligns a fused embedding of the source image and target pose with the target image embedding, it does not explicitly enforce a separation between source and target image representations in the latent space, as illustrated in Fig.~\ref{fig:fig5}(A2). As a result, source appearance information can dominate the fused representation, particularly under large pose variations.

To address this limitation, the proposed \textit{Source-Enhanced Pose Fusion Approach} introduces an explicit mechanism to distinguish source and target image embeddings under the target pose condition. The key idea is to incorporate the source image embedding directly into the contrastive learning objective as an explicit negative sample.

During training, each batch is constructed from identity-preserving triplets $\{x_s^i, x_{tp}^i, x_t^i\}_{i=1}^{B}$. Using the shared IPF module, the source embedding $e_s^i$, the image–pose fused embedding $e_f^i$, and the target image embedding $e_t^i$ are extracted from each triplet. To explicitly introduce source embeddings as negatives, we extend the contrastive batch by reusing the source embeddings, resulting in $2B$ contrastive entries. For the first $B$ entries, the fused image-pose embeddings $e_f^i$ are paired with their corresponding target image embeddings $e_t^i$ as positive pairs. For the remaining $B$ entries, the source embeddings are reused to form additional contrastive entries, thereby explicitly exposing source representations to the contrastive objective.

Under this formulation, each fused embedding $e_f^i$ treats its corresponding source embedding as an explicit negative sample, while all other entries in the batch act as implicit negatives through the InfoNCE normalization. This design enforces a pose-aware separation between source and target representations, encouraging the fused embedding to capture pose-conditioned distinctions rather than collapsing toward source appearance alone. Based on this formulation, we define the pose condition $\psi_p^i$ and the target condition $\psi_t^i$ as follows:


\begin{equation}
    \psi_p^i =
    \left\{
    \begin{array}{ll}
    e_f^i  & \text{if } 1 \leq i \leq B, \\          
    e_{s}^{i-B} & \text{if } B < i \leq 2B.        
    \end{array}
    \right.
    \label{eq:eq4}
\end{equation}

\begin{equation}
    \psi_t^i =
    \left\{
    \begin{array}{ll}
    e_t^i & \text{if } 1 \leq i \leq B, \\
    e_{s}^{i-B} & \text{if } B < i \leq 2B.
    \end{array}
    \right.
    \label{eq:eq5}
\end{equation}

The resulting batch-augmented InfoNCE loss is defined as:
\begin{equation}
    \mathcal{L}_{\text{contr}}
    = \frac{1}{2B} \sum_{i=1}^{2B} 
    \left[
      -\log
      \frac{
          \exp\big(\tau \cdot d(\psi_p^i, \psi_t^i)\big)
      }{
          \sum_{j=1}^{2B} \exp\big(\tau \cdot d(\psi_p^i, \psi_t^j)\big)
      }
      \right],
    \label{eq:eq6}
\end{equation}

where $d(\cdot,\cdot)$ denotes cosine similarity, $\tau$ is a temperature parameter controlling the sharpness of the logits, and $B$ is the batch size. 

By explicitly introducing source embeddings as negatives, Eq.~(6) effectively doubles the number of embedding pairs considered in the contrastive objective and enforces a structured separation between source and target representations. From an objective-function perspective, without explicitly separating source and target embeddings, the contrastive objective admits degenerate solutions in which the fused embedding minimizes loss by collapsing toward dominant source appearance features, especially under large pose discrepancies. Through this explicit negative construction, Eq. (6) reshapes the optimization landscape to penalize such collapse, enforcing a margin that preserves pose-conditioned discrimination between source and target representations.

The effectiveness of this design is validated in the ablation study, where explicitly incorporating source embeddings improves robustness against appearance–pose variations. We note that this formulation is specifically designed for pose-guided synthesis tasks with asymmetric conditioning and may not directly generalize to unconstrained video generation scenarios that require explicit temporal modeling.

\subsection{Fusion Embedding Conditioned Diffusion Model}

\begin{figure*}[t]
\centering
\includegraphics[width=0.9\textwidth]{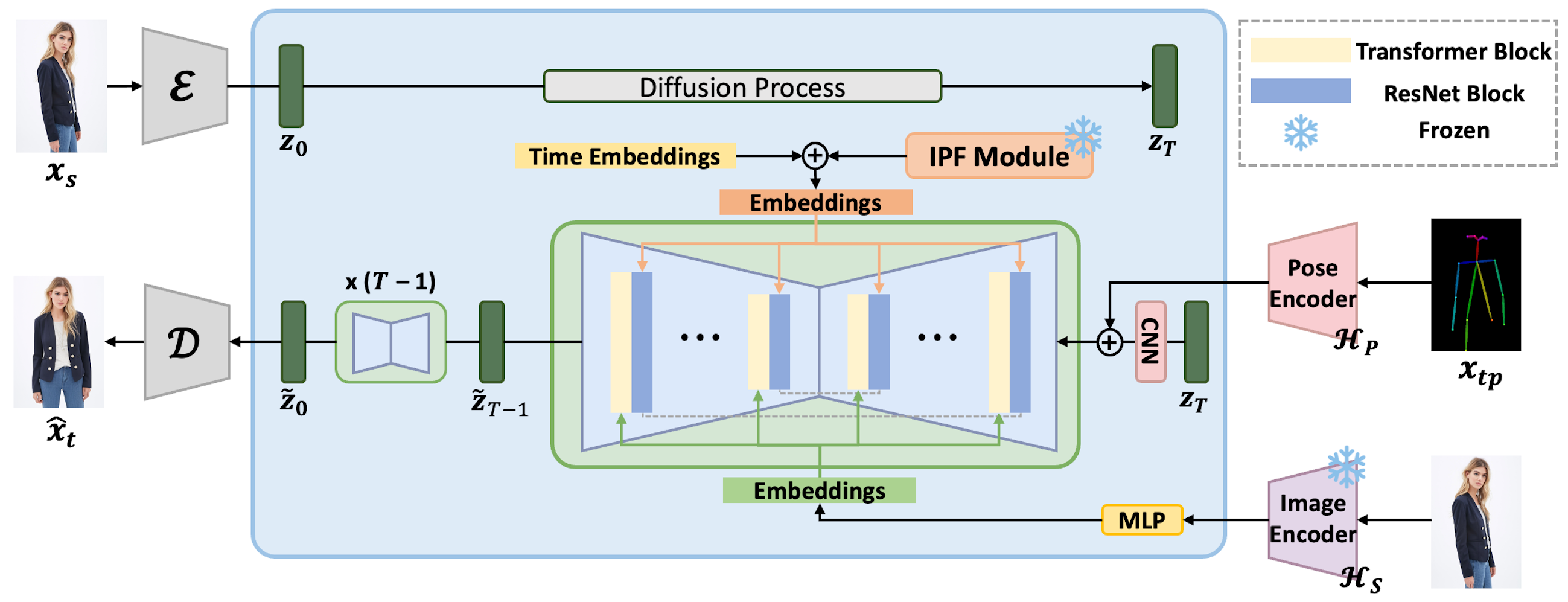} 
\caption{Overall architecture of the proposed FPDM (Fusion Embedding for PGPIS using a Diffusion Model).}

\label{fig:fig2}
\end{figure*}

The proposed FPDM model is built upon the Stable Diffusion (SD) framework. It consists of three main components: a pose encoder $\mathcal{H}_P$, a source image encoder $\mathcal{H}_S$, and an Image–Pose Fusion (IPF) module, which is trained prior to diffusion and used as an explicit conditioning module during the denoising process. The source image encoder $\mathcal{H}_S$ is implemented using a Vision Transformer (ViT), while the pose encoder $\mathcal{H}_P$ adopts a four-layer convolutional architecture from ControlNet~\cite{RN51}. Unlike prior studies that encode each skeletal joint or connection as a separate channel, we employ a standard three-channel RGB representation for pose images. We empirically verify that this representation is sufficient to generate pose-aligned images while maintaining architectural simplicity.

Figure~\ref{fig:fig2} illustrates the overall architecture of the proposed FPDM model. During training, the source image encoder $\mathcal{H}_S$ and the IPF module are kept frozen to preserve stable appearance and fusion representations. Given a noisy latent $z_t$ at timestep $t$, the target pose embedding $e'_{tp}$ is first processed by a single convolutional layer and then concatenated with $z_t$ before being provided to the UNet denoising network $\epsilon_\theta(z_t, t, e'_s, e'_{tp}, e_f)$. The source image embedding $e'_s$, extracted from the source image $x_s$ via $\mathcal{H}_S$, is injected into the transformer blocks of the UNet as key and value features to guide appearance preservation. While the source image embedding provides detailed appearance cues, the fusion embedding $e_f$ captures pose-conditioned appearance relationships learned through contrastive alignment, thereby providing complementary and non-redundant conditioning information. The fusion embedding is combined with the timestep embedding and injected into the ResNet blocks, enabling pose-aware conditioning throughout the denoising process. The training objective is to minimize the mean squared error (MSE) between the ground-truth noise $\epsilon$ and the predicted noise $\epsilon_\theta$. Accordingly, the overall loss function is defined as the expected MSE between the true and predicted noise terms, as formulated below:

\begin{equation}
    \mathcal{L}_{\text{mse}} = \mathbb{E}_{z_0,e'_s,e'_{tp},e_f,\epsilon,t}\left[ \left\| \epsilon - \epsilon_{\theta} (z_t, t, e'_s,e'_{tp},e_f) \right\|_2^2 \right],
    \label{eq:eq7}
\end{equation}

where $z_t$ denotes the noisy latent obtained by adding Gaussian noise $\epsilon$ to the clean latent $z_0$ at timestep $t$. The denoising network $\epsilon_{\theta}$ is conditioned on three inputs: the source image embedding $e'_s$, the target pose embedding $e'_{tp}$, and the fusion embedding $e_f$.

During inference, we adopt a straightforward extension of the standard Classifier-Free Guidance (CFG) scheme to combine multiple conditioning signals by aggregating denoising predictions obtained under conditioning settings. Specifically, we compute three predictions: (i) an unconditional prediction, (ii) a partially conditioned prediction using the source image and target pose, and (iii) a fully conditioned prediction that additionally incorporates the fusion embedding. The final denoising prediction is obtained by a weighted combination of these terms, allowing independent control over the contribution of each conditioning signal. The resulting formulation is given by:

\begin{eqnarray}
\label{eq:eq8}
\epsilon_t = \epsilon_{\theta} (z_t, t, \phi, \phi, \phi) + w_c \cdot \epsilon_{\theta} (z_t, t, e'_s, e'_{tp}, \phi) \nonumber \\
\quad + w_f \cdot \epsilon_{\theta} (z_t, t, e'_s, e'_{tp}, e_f),
\end{eqnarray}

Here, $\phi$ denotes a null (unconditioned) input, while $w_c$ and $w_f$ are guidance weights that control the influence of the source–pose conditioning and the fusion embedding, respectively. 

\section{Experiments}
\subsection{Datasets}
The proposed model is evaluated on two benchmark datasets: DeepFashion In-Shop Clothes Retrieval (DeepFashion) \cite{RN53} and RWTH-PHOENIX-Weather 2014T (Phoenix) \cite{RN55}. DeepFashion consists of 52,712 high-resolution fashion images, and we follow the data configuration described in \cite{RN33}. The model is evaluated at resolutions of 256×176 and 512×352, with input images resized to 256×256 and 512×512, respectively. Phoenix is a German sign language benchmark dataset comprising 7,738 videos recorded from nine signers wearing dark clothing against a gray background, with an image resolution of 210×260 pixels. Following the original dataset configuration, we use 7,096 videos for training and 642 for testing. Pose images are generated by converting the pose coordinates extracted using HRNet \cite{RN58}, following the protocol in \cite{RN57}. To construct the training data, we randomly extract triplets of {source image, target pose, target image} from the video sequences, generating a total of 100,000 pairs. For testing, the first frame of each video serves as the source image, while the subsequent frames provide the target pose sequence, yielding approximately 64,627 test pairs.

\subsection{Metrics}
We evaluate FPDM using a set of widely adopted quantitative metrics. For the DeepFashion dataset, we adopt the Structural Similarity Index Measure (SSIM) \cite{RN59}, Peak Signal-to-Noise Ratio (PSNR), Learned Perceptual Image Patch Similarity (LPIPS) \cite{RN60}, and Fréchet Inception Distance (FID) \cite{RN61}. For the Phoenix dataset, we additionally report Hand SSIM and Hand Pose Error (Hand PE) \cite{RN64}, which are crucial for assessing the quality of hand generation in sign language synthesis. These metrics compare the generated images with the ground truth within cropped hand regions, defined as 60×60 patches centered at the ground-truth middle knuckle. Hand SSIM measures local structural similarity, while Hand PE computes the L1 distance between predicted and ground-truth 2D hand keypoints. The hand keypoints are extracted using HRNetv2 \cite{RN58}. We additionally report the Fréchet Video Distance (FVD) using the evaluation code released by the authors of SignViP \cite{RN214}. For FVD computation, we use a clip length of 30 frames, with clips sampled from random temporal positions within each video. Together, these metrics comprehensively evaluate overall image quality, perceptual fidelity, and fine-grained hand synthesis quality, which are particularly important for sign language generation.

\subsection{Implementations}
All experiments were conducted on a single NVIDIA A100 GPU with 80GB of memory. Our method was implemented using Python 3.9 and PyTorch 2.4 \cite{RN66}. In the prior stage, we adopted the HuggingFace CLIP-Large model as both the image and pose encoders, where the two encoders shared weights to reduce memory consumption. The network was trained for 5 epochs using the AdamW optimizer with a fixed learning rate of $1\times10^{-4}$ and a weight decay of 0.0001. 

For the generative stage, we employed HuggingFace Stable Diffusion 2.1 \cite{RN50} as the base model, using a DINOv2-Large encoder for source image features with an input resolution of $512 \times 512$. The learning rate was initialized at $1\times10^{-4}$ with a linear warm-up during the first 5 epochs, followed by a decay by a factor of 0.1 after 50 epochs. The model was trained for 100 epochs, and training was extended by 10 additional epochs. During training, $e_s$, $e_{tp}$, and $e_f$ were randomly dropped with a probability of $\eta = 20\%$ to enable classifier-free guidance. Following \cite{RN33,RN25}, we trained the DeepFashion model using images of size $512 \times 352$ and evaluated it at the same resolution. For evaluation at $256 \times 176$, an additional model was trained for 10 epochs using $256 \times 256$ source images. For the Phoenix dataset, the model was trained for 20 epochs with a generation resolution of $512 \times 512$, and the generated outputs were resized to $260 \times 210$ pixels for evaluation.

During inference, FPDM follows the standard diffusion sampling procedure and uses the same number of denoising steps as the baseline diffusion models. Inference efficiency was measured on a system equipped with an AMD Ryzen 9 5900X CPU and an NVIDIA RTX 3090 GPU. Under this setting, the diffusion model without the IPF module (ablation B3) requires approximately 2.04 seconds per image, with a peak GPU memory usage of about 17.8 GB. By contrast, the diffusion model with the proposed Image-Pose Fusion (IPF) module (ablation B6) requires approximately 2.95 seconds to generate a single image, with a peak GPU memory consumption of about 21.2 GB. Despite this modest increase, the IPF module introduces no iterative computation during inference and operates on compact latent embeddings, resulting in only a limited additional computational and memory overhead.

\subsection{Quantitative and Qualitative Results}

\begin{table}[!htbp]
\centering
\setlength{\tabcolsep}{2pt}{
\begin{tabular}{lcccccc}
\toprule
Method  & Venue &   FID$\downarrow$    & $\text{FID}_{\text{t}}$$\downarrow$ & LPIPS$\downarrow$  & SSIM$\uparrow$   & PSNR$\uparrow$   \\ \hline
\rowcolor{lightgray} \multicolumn{7}{l}{\textit{Evaluation on 256 $\times$\ 176 \ resolution}}              \\
SPGNet\cite{RN97} & CVPR 21' & 16.184 & 14.107    & 0.2256 & 0.6965 & 17.222 \\
DPTN\cite{RN24} & CVPR 22'   & 17.419 & 15.491    & 0.2093 & 0.6975 & 17.811 \\
NTED\cite{RN25} & CVPR 22' & 8.517  & 6.935     & 0.1770 & 0.7156 & 17.740 \\
CASD\cite{RN27} & ECCV 22'    & 13.137 & 11.619    & 0.1781 & 0.7224 & 17.880 \\
PIDM\cite{RN30} & CVPR 23'   & \underline{6.812}  & \textbf{5.168}     & 0.1867 & 0.6765 & 16.185 \\
PoCoLD\cite{RN38} & ICCV 23'   & 8.067  & -     & 0.1642 & 0.7310 & - \\
CFLD\cite{RN33} & CVPR 24'    & \textbf{6.804}  & 5.688     & 0.1519 & 0.7378 & 18.235 \\
PCDM$^{\dagger}$ \cite{RN36} & ICLR 24'    & 7.699  & 5.901     & 0.1572 & 0.7280 & 18.385 \\
XMDPT-L$^{\ddagger}$\cite{RN93}  & ICML 24'  & 7.438  & 5.579  & 0.1704 & 0.7090 & 17.600 \\
MCLD$^{\dagger}$\cite{RN209} & CVPR 25'  & 6.923 & 5.693  & \underline{0.1482} & \textbf{0.7511} & \textbf{18.840} \\
\textbf{FPDM (Ours)}         &         & 7.318 & \underline{5.459}     & \textbf{0.1445} & \underline{0.7417} & \underline{18.832} \\
\textcolor{lightgray}{VAE Reconstructed}    &        & \textcolor{lightgray}{8.338}  & \textcolor{lightgray}{1.250}     & \textcolor{lightgray}{0.0103} & \textcolor{lightgray}{0.9634} & \textcolor{lightgray}{34.878} \\
\textcolor{lightgray}{Ground Truth}     &            & \textcolor{lightgray}{7.847}  & \textcolor{lightgray}{0.000}     & \textcolor{lightgray}{0.0000} & \textcolor{lightgray}{1.0000} & \textcolor{lightgray}{$+\infty$}       \\ \hline
\rowcolor{lightgray} \multicolumn{7}{l}{\textit{Evaluation on 512 $\times$\ 352 \ resolution}}              \\
CosCosNet2\cite{RN20} & CVPR 21'      & 13.325  & -     & 0.2265 & 0.7236 & - \\
NTED\cite{RN25} & CVPR 22'      & 7.645  & 6.602     & 0.1999 & 0.7359 & 17.385 \\
PoCoLD\cite{RN38} & ICCV 23'   & 8.416  & -     & 0.1920 & 0.7430 & - \\
CFLD\cite{RN33} & CVPR 24'    & \textbf{7.149}  & 6.177     & 0.1819 & 0.7478 & 17.645 \\
PCDM$^{\dagger}$ \cite{RN36} & ICLR 24'      & 7.747  & \underline{6.086}     & 0.1729 & 0.7471 & 18.028 \\
MCLD$^{\dagger}$\cite{RN209} & CVPR 25'  & \underline{7.295} & 6.238  & \underline{0.1757} & \textbf{0.7557} & \textbf{18.211} \\
\textbf{FPDM (Ours)}           &   & 7.534  & \textbf{5.884}     & \textbf{0.1717} & \underline{0.7487} & \underline{18.197} \\
\textcolor{lightgray}{VAE Reconstructed}     &       & \textcolor{lightgray}{8.492}  & \textcolor{lightgray}{1.488}     & \textcolor{lightgray}{0.0212} & \textcolor{lightgray}{0.9149} & \textcolor{lightgray}{30.633} \\
\textcolor{lightgray}{Ground Truth}        &         & \textcolor{lightgray}{8.010}  & \textcolor{lightgray}{0.000}     & \textcolor{lightgray}{0.0000} & \textcolor{lightgray}{1.0000} & \textcolor{lightgray}{$+\infty$}       \\ \bottomrule
\end{tabular}%
}
\caption{
Quantitative comparisons on the DeepFashion dataset were conducted following the evaluation protocol proposed by NTED\cite{RN25}. Only state-of-the-art (SOTA) models with publicly available generation outputs or pretrained weights were included in the comparison. $\text{FID}_{\mathrm{t}}$ denotes the Fréchet Inception Distance (FID) computed between the generated images and the test set only. We strictly followed NTED's evaluation protocol for computing FID, LPIPS, SSIM, and PSNR. $\dagger$ indicates results obtained by evaluating test images generated directly by the authors. $\ddagger$ denotes results obtained using test images generated by the publicly released model provided by the authors. The remaining results were referenced from experiments reported in CFLD\cite{RN33}. Bold and underlined values represent the best and second-best performance, respectively.}
\label{table:t1}
\end{table}

The proposed model was evaluated both quantitatively and qualitatively using the DeepFashion dataset and compared with SPGNet \cite{RN97}, DPTN \cite{RN24}, NTED \cite{RN25}, CASD \cite{RN27}, PIDM \cite{RN30}, CFLD \cite{RN33}, PCDM \cite{RN36}, XMDPT \cite{RN93} and MCLD \cite{RN209}, all of which provide either publicly available generated results or pretrained weights. In addition, PoCoLD \cite{RN38} was included for quantitative comparison only, since its official evaluation code is publicly available and follows the same evaluation protocol as NTED, although the pretrained model weights and the generated test images are not publicly released. 

Quantitative evaluation was conducted using FID, FID$_\mathrm{t}$, LPIPS, SSIM, and PSNR. As noted by Lu et al. (2024) ~\cite{RN33}, fair comparisons are difficult because different methods employ different parameter settings for evaluation metrics. Following their protocol, we strictly adhered to NTED’s evaluation procedure for computing FID, LPIPS, SSIM, and PSNR. Furthermore, as discussed by Han et al.~\cite{RN38}, the DeepFashion dataset exhibits inherent limitations for conventional FID-based evaluation due to distributional discrepancies between the training and test sets. As shown in Table~\ref{table:t1}, the FID score between the training and test sets reaches 7.847 at a resolution of 256×176, indicating a significant distribution gap. To provide a more reliable assessment, we therefore additionally report FID$_\mathrm{t}$, computed between the generated images and the test set.

\begin{figure}[t]
\centering
\includegraphics[width=1.0\columnwidth]{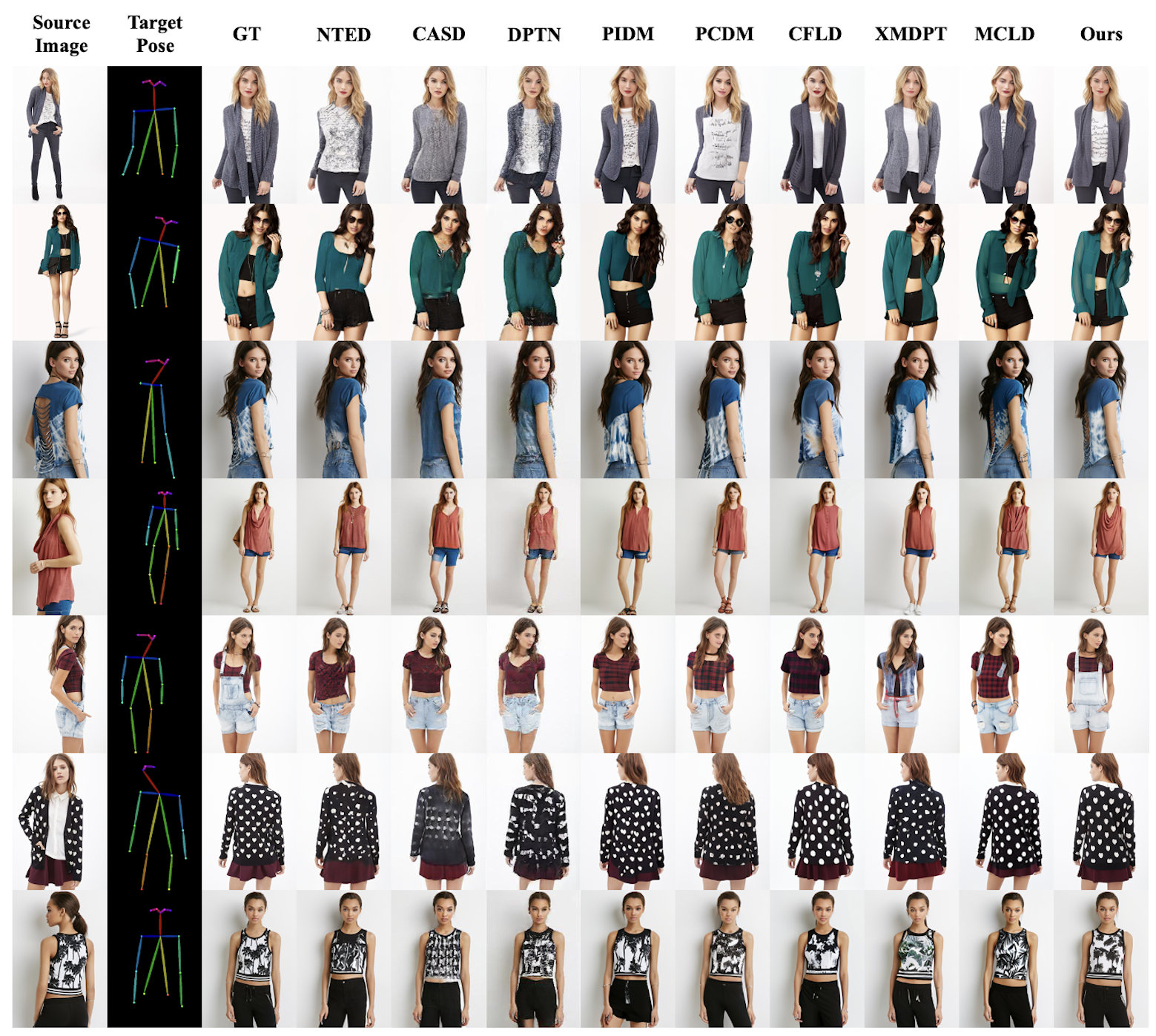} 
\caption{Qualitative comparisons with current state-of-the-art models on the DeepFashion dataset.}
\label{fig:fig3}
\end{figure}

As shown in Table~\ref{table:t1}, FPDM demonstrates consistently strong performance across complementary evaluation metrics, reflecting a balanced trade-off between perceptual realism, structural consistency, and identity preservation. At the 256×176 resolution, although some methods achieve lower training-set FID scores, such improvements do not necessarily translate to better generalization on the test set. For example, CFLD attains a low training FID but exhibits a noticeable increase in FID$_\mathrm{t}$, suggesting potential overfitting to the training distribution. In contrast, FPDM achieves the second-lowest FID$_\mathrm{t}$ and the best LPIPS score, indicating superior perceptual similarity and texture preservation. It also ranks second in SSIM and PSNR, confirming accurate structural alignment and pixel-level fidelity.

A similar trend is observed at the higher 512×352 resolution. FPDM consistently achieves the best LPIPS and ranks within the top two for FID$_\mathrm{t}$, SSIM, and PSNR. These results indicate that FPDM maintains robust appearance preservation and pose-consistent synthesis across resolutions, aligning well with the design objective of the proposed source-aware fusion embedding. Rather than optimizing a single metric in isolation, FPDM achieves a balanced performance across perceptual and structural criteria that are particularly relevant for pose-guided person image synthesis.

Figure~\ref{fig:fig3} presents a visual comparison of the proposed model with recent SOTA methods on the DeepFashion dataset. FPDM shows improved performance in generating high‑quality target images by more accurately capturing fine‑grained visual details corresponding to pose variations. Rows 1–3 illustrate cases where the target images share the same orientation as the source images; in these scenarios, FPDM effectively reproduces the detailed appearance of the source. For instance, the sweater and T‑shirts in Row 1, the blouse styling in Row 2, and the shredded back of the TeeTank in Row 3 are all faithfully reconstructed. Rows 4 and 5 highlight situations where the source image reveals only partial information about the target image. In Row 4, FPDM reconstructs the full silhouette of a draped‑neckline blouse from its visible left‑side outline, while in Row 5, it synthesizes the entire jumpsuit from the partially visible source region. Rows 6 and 7 demonstrate cases where the generated areas are completely occluded in the source. FPDM naturally transfers the white heart‑pattern detail from the front to the back of the blouse in Row 6, and convincingly generates the front view of a palm‑tree pattern in Row 7 based on its back appearance. These results indicate that FPDM can infer and synthesize structurally consistent and stylistically accurate representations, even when the source image provides only limited or partial visual cues.

\begin{figure}[t]
\centering
\includegraphics[width=1.0\columnwidth]{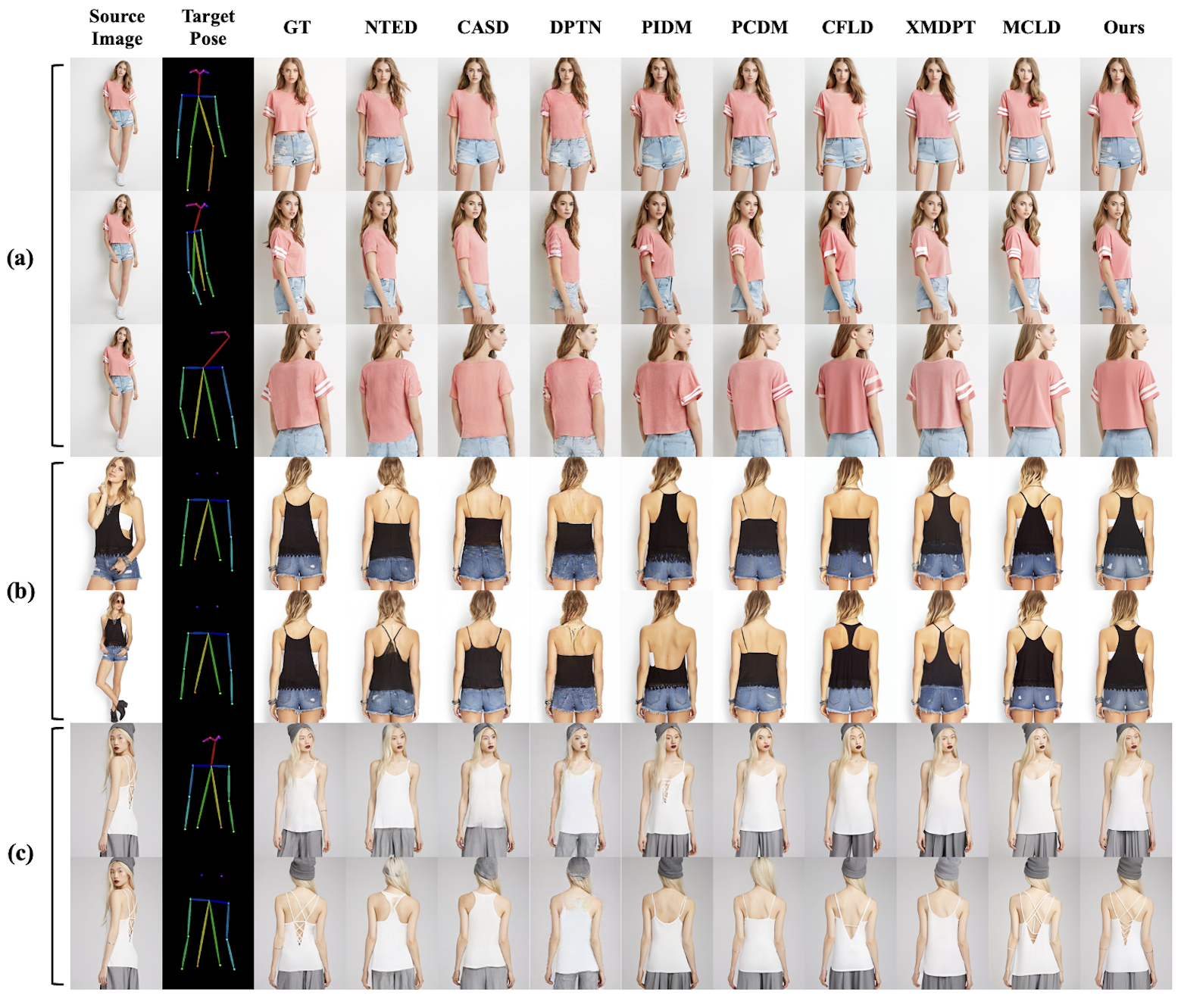} 
\caption{Qualitative visualization of FPDM's robustness to pose and appearance variations on the DeepFashion dataset.}
\label{fig:fig4}
\end{figure}

As shown in Figure~\ref{fig:fig4}, FPDM consistently generates visually coherent images under variations in both pose and source appearance. Figure~\ref{fig:fig4}(a) demonstrates that when different target poses are applied to the same source image, the generated images faithfully preserve key stylistic elements. In particular, FPDM consistently reproduces the two‑stripe sleeves on both arms with the correct position and shape. 

By contrast, NTED, CASD, DPTN, and PIDM fail to reproduce this detail accurately. While CFLD, PCDM, XMDPT, and MCLD can generate the stripe, they fail to maintain its positional and shape consistency. Figure~\ref{fig:fig4}(b) illustrates the scenario where the target pose is fixed while the source pose varies. FPDM continues to generate stable and consistent results, accurately reproducing the TeeTank top as well as the inner Tube Top, whereas the other SOTA models tend to lose consistency in upper‑garment details. Figure~\ref{fig:fig4}(c) evaluates front‑to‑back consistency when generating from a single side‑view source image. FPDM successfully maintains visual continuity between the front and back, accurately reflecting the shoulder straps and back design of the Double Spaghetti Strap Top visible from the side view. These results demonstrate that FPDM generates high-quality, pose-conditioned images while preserving strong visual consistency and structural integrity.

\subsection{Ablation Study}
\subsubsection{Image-Pose Fusion Module}

\begin{figure}[t]
\centering
\includegraphics[width=0.6\columnwidth]{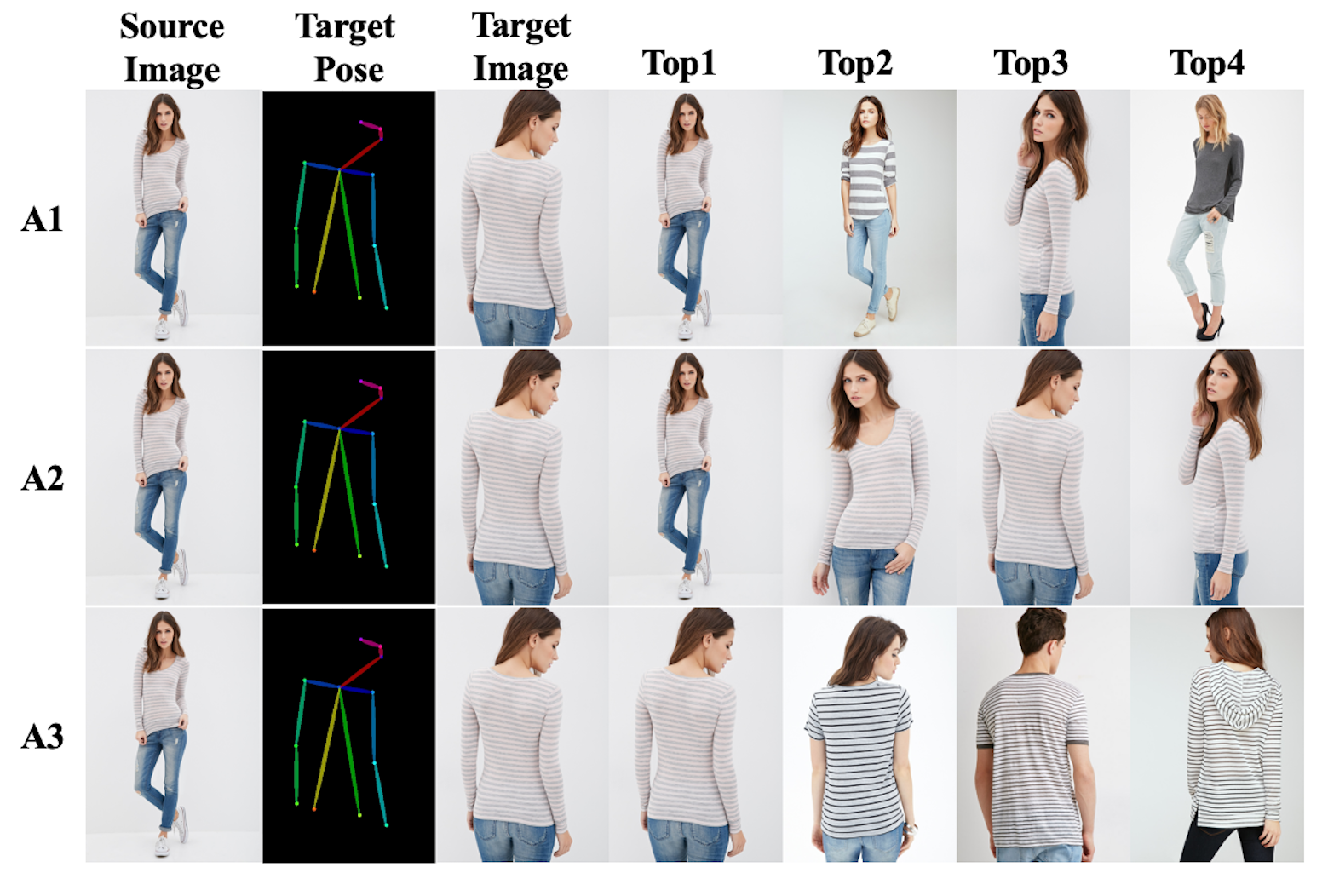} 
\caption{Qualitative evaluation of Image-Pose Fusion (IPF) module ablation. Cosine similarity–based ranking results for three IPF configurations (A1–A3) on the test set. A3 achieves superior alignment in both pose and appearance.}
\label{fig:fig5}
\end{figure}

The Image-Pose Fusion (IPF) module in this study aims to generate an embedding aligned with the target image representation the target image by integrating the source image and the target pose. To evaluate its performance under different configurations, we designed three experimental settings: A1, A2, and A3. In A1, the source image is directly input into a pre-trained CLIP image encoder to extract its embedding. A2 follows the baseline fusion approach described in Equation~\ref{eq:eq3}, while A3 adopts the proposed Source-Enhanced Pose Fusion strategy described in Equation~\ref{eq:eq6}. In A1, the embedding is obtained solely from the source image, whereas A2 and A3 compute embeddings by fusing the source image and target pose. To assess alignment quality, we compute cosine similarity between each fusion embedding and all target image embeddings in the test set. The quantitative results for these settings are reported in Table~\ref{table:t2}, with qualitative examples presented in Figure~\ref{fig:fig5}.

\begin{table}[!htbp]
\centering
\setlength{\tabcolsep}{1mm}{%
\begin{tabular}{ccccccc}
\toprule
Method & R@1            & R@3            & R@5            & Average rank     \\ \hline
Random  & 0.000          & 0.001          & 0.002          & 1597            \\
A1      & 0.000          & 0.450          & 0.609          & 63.85           \\
A2      & 0.050          & 0.793          & 0.976          & 3.474           \\
A3      & \textbf{0.997} & \textbf{0.985} & \textbf{0.998} & \textbf{1.325}  \\ \bottomrule
\end{tabular}%
}
\caption{Qualitative results of the Image-Pose Fusion (IPF) module ablation study. We compare retrieval results based on cosine similarity between the query image embedding and all test image embeddings. Target images are ranked according to similarity, and retrieval performance is summarized using Recall@k. The results demonstrate the impact of the IPF module on embedding quality and retrieval accuracy.}
\label{table:t2}
\end{table}

Table~\ref{table:t2} reports the quantitative results of target image retrieval based on cosine similarity, comparing the embeddings produced by A1–A3 with a random baseline. As expected, random retrieval yields an extremely low probability of correctly identifying the target image. A1, which relies solely on the pre-trained CLIP embedding of the source image, achieves a Recall@5 of 0.609 but a near-zero Recall@1, highlighting its limitation in precise alignment. A2, which employs the baseline IPF approach, significantly improves Recall@5 to 0.976. However, its Recall@1 remains notably low at 0.050, suggesting that the fused embedding is still not precisely aligned with the target image. In contrast, A3, incorporating the proposed Source-Enhanced Pose Fusion strategy, achieves the best performance with a Recall@5 of 0.998 and a Recall@1 of 0.997 (99$\%$), while maintaining an exceptionally low average rank of 1.325. These results quantitatively demonstrate that the Source-Enhanced Pose Fusion approach enables the IPF module to effectively integrate source appearance and target pose information. This synergy produces highly accurate embeddings that are precisely aligned for target image retrieval.

Figure~\ref{fig:fig5} presents a visual comparison of the three Image-Pose Fusion (IPF) configurations (A1–A3). A1, which uses only source image embeddings from a pretrained CLIP encoder, retrieves visually similar candidates (e.g., striped tops in the Top‑1 to Top‑3 results) but fails to capture pose‑specific attributes, such as the target’s back‑view orientation. This indicates that CLIP embeddings are sensitive to general appearance features (e.g., clothing style, gender, posture) but are insufficient for encoding fine‑grained pose information. A2, which implements the baseline IPF approach, retrieves garments that are visually similar to the source but still show poor structural alignment with the target pose. In contrast, A3, which adopts the proposed Source‑Enhanced Pose Fusion strategy, consistently retrieves the correct target image at Top‑1 and maintains high retrieval accuracy for pose‑ and texture‑consistent candidates up to Top‑4. These results demonstrate that A3 effectively integrates appearance and pose information, producing embeddings that are both semantically richer and structurally more precise.

\subsubsection{Fusion Embedding Conditioned Diffusion Model.} 

\begin{table}[!htbp]
\centering
\setlength{\tabcolsep}{1mm}{%
\begin{tabular}{cccccc}
\toprule
Method & FID$\downarrow$            & $\text{FID}_{\text{t}}$$\downarrow$            & LPIPS$\downarrow$          & SSIM$\uparrow$            & PSNR$\uparrow$             \\ \hline
B1     & 7.607          & 6.243          & 0.1824          & 0.7443          & 17.768           \\
B2     & \textbf{7.349}          & 6.073          & 0.1885          & 0.7379          & 17.627           \\
B3     & 7.506          & 5.912          & 0.1737          & 0.7470          & 18.1254          \\
B4     & \underline{7.464} & \textbf{5.797} & \underline{0.1734}    & 0.7466    & \underline{18.1315}    \\
B5     & 7.596          & 6.0346         & 0.1737          & \underline{0.7475}          & 18.1233          \\
B6     & 7.534    & \underline{5.884}    & \textbf{0.1717} & \textbf{0.7487} & \textbf{18.1969} \\ \bottomrule
\end{tabular}%
}
\caption{Quantitative comparison of generation model ablation results. All models are evaluated at a resolution of 352×512 pixels. B6 shows the best overall performance, validating the effectiveness of the proposed Source-Enhanced Fusion strategy.}
\label{table:t3}
\end{table}

We define six configurations of the proposed generative model (B1–B6) and evaluate them through comparative experiments. B1–B3 exclude the IPF module and differ only in the choice of source image encoder and input resolution. Specifically, B1 employs the CLIP‑Large model as the source image encoder with an input resolution of 224, B2 uses the DINOv2‑Large encoder at the same resolution, and B3 adopts DINOv2‑Large with an increased input resolution of 512. In contrast, B4–B6 integrate the IPF module into the B3 backbone, with each configuration corresponding to one of the IPF settings (A1–A3) introduced in the Image‑Pose Fusion Module subsection. B4 uses a simple IPF configuration in which a CLIP encoder extracts the source image embedding for the generative model. B5 incorporates an IPF module trained with the baseline approach (Equation~\ref{eq:eq3}), whereas B6 employs an IPF module trained with the proposed Source‑Enhanced Pose Fusion strategy (Equation~\ref{eq:eq6}).

Table~\ref{table:t3} reports the quantitative results of the ablation study for configurations B1–B6. Comparing B1 and B2, B1—which uses the CLIP encoder with an input resolution of 224—achieves slightly better LPIPS, SSIM, and PSNR scores than B2, which employs the DINOv2 encoder at the same resolution. Increasing the input resolution of the DINOv2 encoder to 512 in B3 leads to a clear improvement across all metrics compared with B2. Incorporating the IPF module further enhances performance. B4, which simply introduces the IPF module with CLIP embeddings, and B5 and B6, which employ the baseline and Source‑Enhanced IPF strategies, respectively, all outperform B3 overall. Among them, B6 achieves the best LPIPS, SSIM, and PSNR scores (excluding FID and $\text{FID}_{\mathrm{t}}$) and ranks second in $\text{FID}_{\mathrm{t}}$, providing quantitative evidence for the effectiveness of the proposed Source-Enhanced Fusion strategy.

\subsubsection{Effect of the Fusion Embedding} 

\begin{figure}[t]
\centering
\includegraphics[width=1.0\columnwidth]{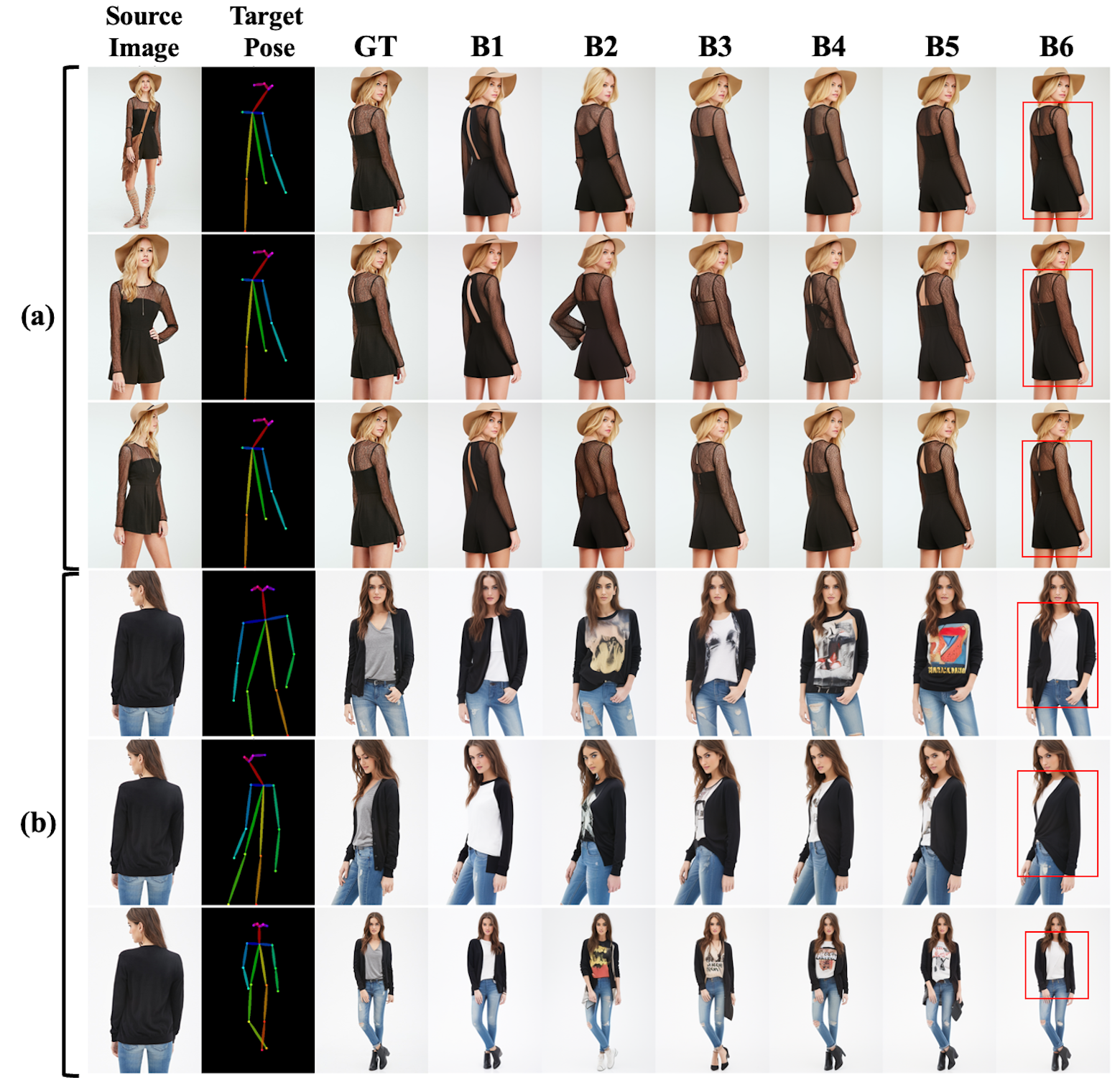} 
\caption{Effect of the Fusion Embedding in the ablation study. Models B1–B6 are compared under diverse pose conditions. B6 shows the most consistent appearance synthesis across variations.}
\label{fig:fig7}
\end{figure}

The results indicate that the proposed B6 configuration consistently preserves coherent appearance under varying input conditions. Figure~\ref{fig:fig7}(a) demonstrates the case where different source images with diverse poses are mapped to the same target pose. The task involves generating a back-view image of a woman wearing a short black dress with a sheer lace upper body from multiple front-view source images. B6 accurately infers the dress structure and texture, producing a consistent back‑view appearance, whereas B1–B3 fail to capture the garment structure, and B4–B5 yield outputs with inconsistent visual appearance. Figure~\ref{fig:fig7}(b) shows the reverse scenario, where a single back‑view source image of a woman in a black cardigan is transformed into multiple front‑view target poses. Since the source lacks visible information about the inner top, the model must infer it. B6 plausibly generates a white inner top and maintains consistent outer‑garment appearance across poses. By contrast, B1–B5 exhibit inconsistent outputs, with the upper garment varying between shirt‑ and sweatshirt‑like appearances. These qualitative observations demonstrate that the Source-Enhanced Fusion Embedding provides robust alignment with the target representation and enables stable, stylistically consistent synthesis across variations in source pose and appearance.

\subsubsection{Convergence Analysis of Image–Pose Fusion Ablation} 

\begin{figure}[t]
\centering
\includegraphics[width=1.0\columnwidth]{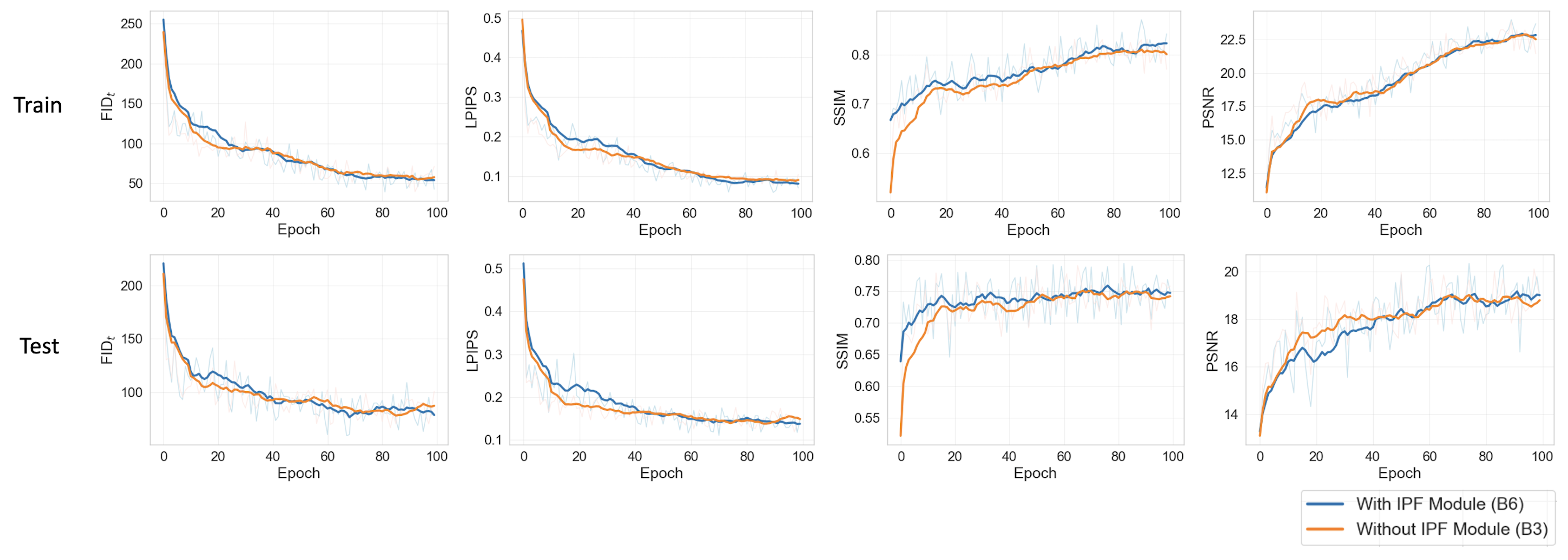}
\caption{Convergence curves of FPDM ablation variants B3 and B6, without and with the Image–Pose Fusion (IPF) module, evaluated on the test set. The plots illustrate the evolution of FID$_t$, LPIPS, SSIM, and PSNR across training epochs, smoothed using a moving average with a window size of 10. At each epoch, metrics are computed using a fixed number of generated samples per variant (20 for B3 and 10 for B6), due to a minor difference in the experimental setup during training. Test-set curves are reported for analysis only and were not used for model selection, early stopping, or hyperparameter tuning.}
\label{fig:fig10}
\end{figure}

Figure~\ref{fig:fig10} compares the convergence behaviors of the B3 and B6 variants, corresponding to FPDM without and with the Image–Pose Fusion (IPF) module. Both ablation variants converge within 100 training epochs. Except for SSIM, the B3 variant exhibits faster convergence in the early training stage across most metrics, suggesting that optimization proceeds more quickly in the absence of the IPF module, albeit in a less structured manner. However, after approximately 40–50 epochs, the performance gap between the two variants narrows, and in the later stages of training, the B6 variant consistently achieves higher performance across most metrics. This trend indicates that while the IPF module may introduce additional optimization complexity during early training, it ultimately contributes to learning more stable and superior representations, particularly in terms of perceptual quality and pose consistency in the final generated results.

\subsection{Applications to Sign Language Video Generation}

Research on sign language using artificial intelligence aims to automate both recognition and generation, thereby enhancing social, educational, and professional accessibility for signers, including the Deaf community. Within this field, sign language video generation has emerged as a prominent topic in computer vision \cite{RN64,RN101,RN203,RN204,RN205,RN214}. Nevertheless, existing approaches often produce unrealistic or low‑quality results, which limits their practical applicability. In this study, we explore real‑world applications such as pose‑to‑sign language image synthesis and signer anonymization for identity protection. To this end, we train the proposed FPDM model on the RWTH‑PHOENIX‑Weather 2014T (PHOENIX) dataset and evaluate its performance through both quantitative and qualitative analyses.

\begin{table}[!htbp]
\centering{%
\begin{tabular}{lcccccc}
\toprule
Method & Venue & FVD$\downarrow$ & FID$\downarrow$ & SSIM$\uparrow$  & Hand SSIM$\uparrow$ & Hand Pose$\downarrow$ \\ \hline
EDN$^{\dagger}$ \cite{RN98} & ICCV 19' & - & 41.54     & 0.737     & 0.553     & 23.09     \\
Vid2vid$^{\dagger}$ \cite{RN99} & NeurIPS 18' & -& 56.17     & 0.750     & 0.570     & 22.51     \\
Pix2PixHD$^{\dagger}$ \cite{RN100} & CVPR 18' & -& 42.57     & 0.737     & 0.553     & 23.06    \\
NMT$^{\dagger}$ \cite{RN101} & IJCV 20' & -& 64.01     & 0.727     & 0.533     & 23.17     \\
SignGAN \cite{RN64} & CVPR 22' & -& 27.75  & 0.759 & 0.605   & 22.05  \\
ControlNet$^{\ddagger}$ \cite{RN212} & CVPR 23' & 556.6 & -  & 0.784 & 0.505   & -  \\
AnimateAnyone$^{\ddagger}$ \cite{RN213} & CVPR 24' & 365.4 & - &  0.794 & 0.505   & -     \\
SignDiff \cite{RN205} & FG 25' & -  & \underline{25.22}  & \underline{0.849} & \underline{0.676}   & \textbf{20.04}     \\
SignVDM \cite{RN214} & NeurIPS 25' & \underline{275.2} & -  & 0.829 & 0.614   & -     \\
\textbf{FPDM (Ours; B6)} &  & \textbf{106.7}& \textbf{5.13}     & \textbf{0.864}    & \textbf{0.682}     & \underline{21.43}     \\ \bottomrule
\end{tabular}%
}
\caption{
Quantitative comparison on the RWTH-PHOENIX-Weather 2014T dataset using sign language–specific evaluation metrics. $^{\dagger}$ and $^{\ddagger}$ indicate results reported by the authors of SignGAN~\cite{RN64} and SignVDM~\cite{RN214}, respectively. Unavailable metrics are marked with ``--'' and bold values denote the best performance.}
\label{table:t4}
\end{table}
\begin{figure}[t]
\centering
\includegraphics[width=1.0\columnwidth]{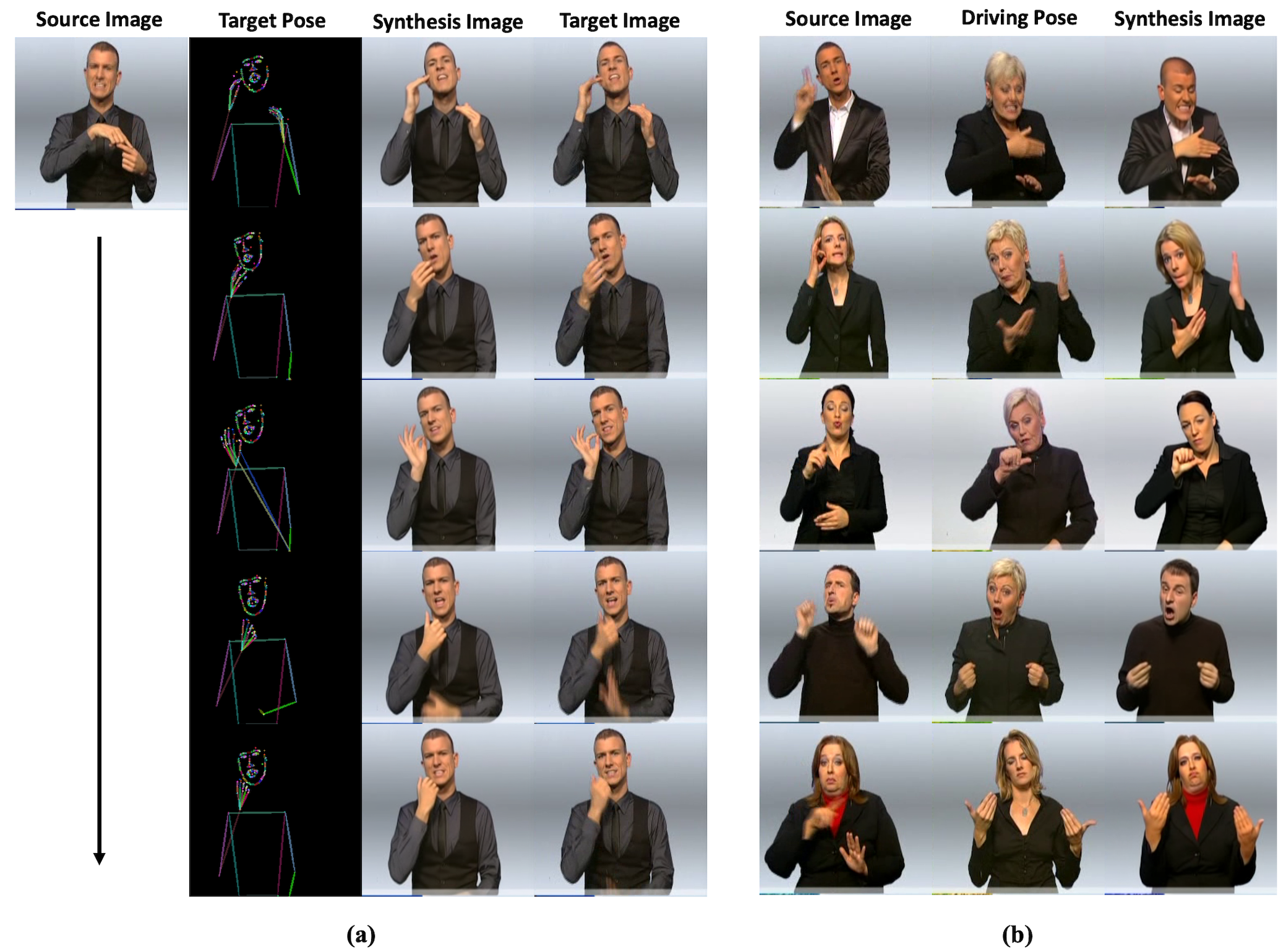} 
\caption{Visualization of results on the RWTH-PHOENIX-Weather 2014T dataset. (a) Signer generation results, where the source image corresponds to the first frame of a test video and the generated images are presented in temporal order according to the driving poses. (b) Signer anonymization results produced by FPDM, where the source image and pose sequence are randomly matched. FPDM generates identity-preserving and pose-consistent images, demonstrating its effectiveness for sign language video generation and its potential for signer de-identification.}
\label{fig:fig8}
\end{figure}

Table~\ref{table:t4} presents a quantitative comparison between the proposed FPDM model (B6) and representative state-of-the-art methods for sign language video generation, including EDN\cite{RN98}, Vid2vid\cite{RN99}, Pix2PixHD\cite{RN100}, NMT\cite{RN101}, SignGAN\cite{RN64}, and the recent diffusion-based approach SignDiff\cite{RN205}, ControlNet\cite{RN212} , AnimateAnyone\cite{RN213}, and the Sign Video Diffusion Model\cite{RN214}, are also included. FPDM achieves an FVD of 106.7, FID of 5.13 and an SSIM of 0.886, demonstrating its ability to generate visually realistic and high-quality sign language images. More importantly, FPDM shows clear advantages in hand-specific evaluation, which is critical for sign language synthesis. It attains the highest Hand SSIM score of 0.682 and a competitive Hand Pose error of 21.43, indicating accurate and consistent hand appearance and motion. These results suggest that FPDM effectively captures both global visual fidelity and fine-grained hand details, outperforming existing methods in terms of sign language–specific metrics. Although FPDM focuses on pose-guided image synthesis rather than end-to-end video generation, its strong performance on hand-centric and structural evaluation metrics indicates its suitability as a core appearance modeling component for sign language video generation pipelines.

Figure~\ref{fig:fig8} presents sampled visualizations of sign language image generation results. Figure~\ref{fig:fig8}(a) illustrates signer generation results. In this example, the appearance of the source image is transferred to a new sequence driven by the pose of a different input image. The proposed model generates images while naturally and consistently preserving the appearance of the source image. In some frames, the hand regions are rendered with comparable or occasionally higher visual clarity than those observed in the ground truth image, demonstrating superior visual quality. This suggests that the proposed model is capable of generating high-quality images based on the given pose, without being overly constrained by the visual quality of individual source or target frames. Figure~\ref{fig:fig8}(b) shows an example of a signer anonymization experiment, where sign language images are generated based on the appearance of a source image corresponding to a given driving image. In this process, the pose of the driving image, extracted using the pre-trained DWpose model, and a randomly selected source image are used as inputs. Although the PHOENIX dataset used for training consists of triplets composed of a source image, target pose, and target image from the same signer, the model generalizes well to mismatched inputs. The generated outputs accurately reflect the driving pose while stably preserving the appearance of the source image. This provides qualitative evidence that the driving pose functions primarily as a guiding signal for pose information, with limited influence on appearance in the generation process. Overall, the qualitative results in Figure~\ref{fig:fig8} demonstrate that the proposed FPDM enables natural and identity-preserving sign language image generation and provides a practical foundation for applications such as signer de-identification and synthetic sign language video generation.

\section{Limitations}

\begin{figure}[t]
\centering
\includegraphics[width=1.0\columnwidth]{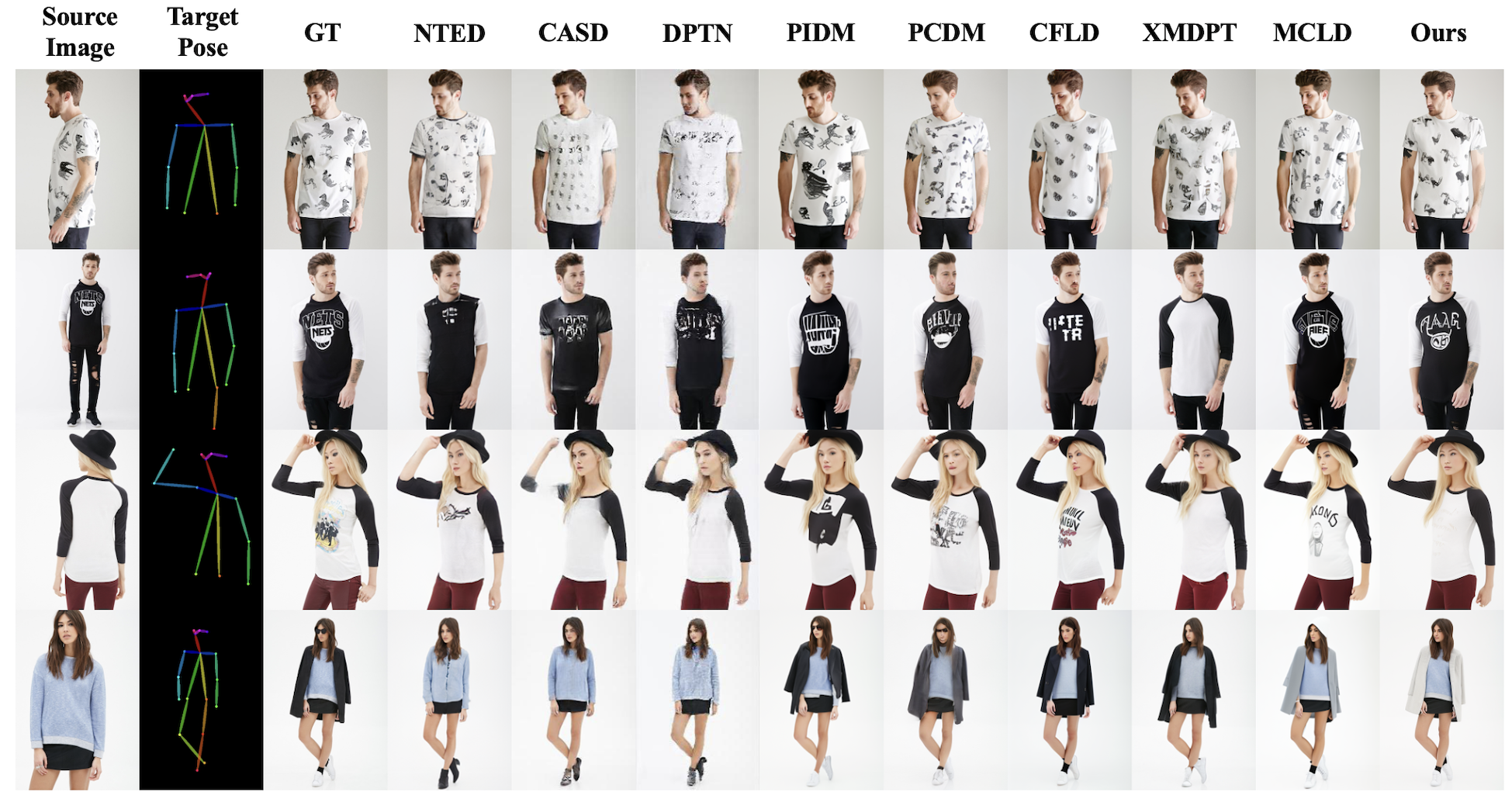} 
\caption{Failure cases of FPDM. Examples include missing fine details (e.g., text, logos), incorrect front-view synthesis, and hallucinated accessories or distorted textures in challenging regions.}
\label{fig:fig9}
\end{figure}

This study has several limitations. First, although recent state-of-the-art methods such as RePoseDM \cite{RN34} and ImagePos \cite{RN208} report high-quality generation results, direct quantitative comparison is challenging. The evaluation metrics used in pose-guided person image synthesis (PGPIS) research vary across studies, and these recent methods often do not release source code or generated outputs, which restricts reproducible evaluation. 

Second, the proposed FPDM still exhibits certain limitations in generation performance. As shown in Figure~\ref{fig:fig9}, FPDM occasionally fails to reproduce fine-grained patterns such as animals, logos, and text (Rows 1–2). In addition, it sometimes generates incorrect front views for unseen regions (Row 3), which can be attributed to probabilistic inference from back-view observations. Moreover, because some full-body images in the DeepFashion dataset include accessories or outerwear, FPDM may occasionally hallucinate such items during full-body synthesis. In addition, the performance differences among the B3–B6 configurations are relatively small under the current evaluation setting, and the model may fail to reproduce fine textures or heavily occluded regions, resulting in blurred or distorted outputs in challenging cases. Moreover, FPDM inherits the computational characteristics of diffusion-based generation models, which require iterative multi-step sampling during inference and thus incur higher computational cost than single-pass generative models. This limitation is intrinsic to the diffusion framework rather than specific to our design.

Future work will focus on improving fine-grained texture fidelity and conducting broader comparisons as additional models and evaluation results become publicly available. Finally, although FPDM is evaluated in the context of sign language video generation, it operates at the image synthesis level and does not explicitly model temporal dynamics. The proposed source-enhanced pose fusion approach is therefore specifically designed for pose-guided synthesis tasks that require strict identity preservation under large pose variations. Extending this formulation to fully general video generation frameworks with temporal modeling remains an interesting direction for future work, but is beyond the scope of this paper.

\section{Conclusion}
In this paper, we propose a novel diffusion-based approach for Pose-Guided Person Image Synthesis (PGPIS). The proposed method, Fusion Embedding for PGPIS with Diffusion Model (FPDM), introduces an Image-Pose Fusion (IPF) module that learns a fusion embedding aligned with the target image via contrastive learning and employs it as conditioning for the Latent Diffusion Model (LDM). This architecture improves robustness to variations in both the source image and the target pose. To evaluate the effectiveness of FPDM, we conduct experiments on the DeepFashion dataset, which reflects real-world e-commerce scenarios, and the PHOENIX dataset, which is used for sign language video generation. Experimental results demonstrate that FPDM achieves competitive performance against existing PGPIS methods in both quantitative and qualitative evaluations. In particular, the proposed Source-Enhanced Pose Fusion enables consistent target image generation for the same identity while stably preserving identity-specific visual characteristics across variations in both the target pose and the source image. However, accurately transferring fine-grained patterns such as text or intricate textures from the source image to the generated image remains a challenge. Future work will explore extensions of FPDM to image synthesis and temporally structured video synthesis tasks that require a deeper understanding of fine-grained contextual cues in source images.

\section*{Acknowledgments}
This work was supported by the National Research Foundation of Korea (NRF) grant funded by the Korea government (MSIT) (No. 2023R1A2C100697011).

\bibliographystyle{unsrt}  
\bibliography{reference}  
\end{document}